\documentclass[twoside,leqno,twocolumn]{article}

\usepackage[letterpaper]{geometry}

\usepackage{ltexpprt}

\usepackage{graphicx}
\usepackage{balance}  
\usepackage{lipsum}
\usepackage{tabularx}
\usepackage{graphicx}
\usepackage{epstopdf}
\usepackage{psfrag}
\usepackage{color}
\usepackage{filecontents}
\usepackage{cite}
\usepackage{mathtools}

\usepackage{amssymb,amsmath,amsfonts}
\let\emptyset\varnothing
\usepackage{float}
\usepackage{algorithm}
\usepackage[noend]{algpseudocode}

\usepackage{tikz}
\usepackage{ifthen}
\usepackage{subfigure}
\usepackage{overpic}
\usepackage{pict2e}

\usepackage{caption}
\usepackage{subfigure,float}
\usepackage{footnote}
\usetikzlibrary{matrix,positioning,decorations.pathreplacing}

\usepackage{changepage}

\newcommand{\algo}{{\sc LExiCoL}}

\newcommand{\cotrain}{{\sc co-training}}
\newcommand{\tptrain}{{\sc tp-training}}
\newcommand{\mltrain}{{\sc ml-training}}

\newcommand{\LMatrix}{\mathbf{L}}
\newcommand{\DMatrix}{\mathbf{D}}

\newcommand{\IMatrix}{\mathbf{I}}
\newcommand{\XMatrix}{\mathbf{X}}
\newcommand{\HMatrix}{\mathbf{H}}
\newcommand{\YMatrix}{\mathbf{Y}}
\newcommand{\PMatrix}{\mathbf{P}}
\newcommand{\RMatrix}{\mathbf{R}}
\newcommand{\network}{\cal{G}}

\newcommand{\nodeset}{\cal{V}}
\newcommand{\edgeset}{\cal{E}}

\newcommand{\AMatrix}{\mathbf{A}}                  

\begin{document}

\title{\Large Expanding Label Sets for Graph Convolutional Networks }
\author{Mustafa Co{\c s}kun\thanks{ Department of Computer Engineering, Abdullah G{\"u}l University, Turkey. Email:
mustafa.coskun@agu.edu.tr}
\and 
Burcu Bakir-Gungor\thanks{ Department of Computer Engineering, Abdullah G{\"u}l University, Turkey. Email:
burcu.gungor@agu.edu.tr}
\and Mehmet Koyut{\"u}rk\thanks{Department of Electrical Engineering \& Computer Science,
Case Western Reserve University, USA. Email: mehmet.koyuturk@case.edu}}

\date{}

\maketitle

\fancyfoot[R]{\scriptsize{Copyright \textcopyright\ 2020 by SIAM\\
Unauthorized reproduction of this article is prohibited}}
In recent years, Graph Convolutional Networks (GCNs) and their variants have been widely utilized in learning tasks that involve graphs. These tasks include recommendation systems, node classification, among many others. In node classification problem, the input is a graph in which the edges represent the association between pairs of nodes, multi-dimensional feature vectors are associated with the nodes, and some of the nodes in the graph have “known” labels. The objective is to predict the labels of the nodes that are not labeled, using the nodes’ features, in conjunction with graph topology. While GCNs have been successfully applied to this problem, the caveats that they inherit from traditional deep learning models pose significant challenges to broad utilization of GCNs in node classification. One such caveat is that training a GCN requires a large number of labeled training instances, which is often not the case in realistic settings. To remedy this requirement, state-of-the-art methods leverage network diffusion-based approaches to propagate labels across the network before training GCNs. However, these approaches ignore the tendency of the network diffusion methods in biasing proximity with centrality, resulting in the propagation of labels to the nodes that are well-connected in the graph.To address this problem, here we present an alternate approach, namely \algo, to extrapolating node labels in GCNs in the following three steps: (i) clustering of the network to identify communities, (ii) use of network diffusion algorithms to quantify the proximity of each node to the communities, thereby obtaining a low-dimensional topological profile for each node, (iii) comparing these topological profiles to identify nodes that are most similar to the labeled nodes. Testing on three large-scale real-world networks that are commonly used in benchmarking GCNs, we systematically evaluate the performance of the proposed algorithm and show that our approach outperforms existing methods for wide ranges of parameter values.

\section{Introduction}
\label{sec:introduction}
Graph Convolutional Networks (GCNs)~\cite{Kipf} are a variant of Convolutional Neural Netwoks (CNNs) in which the underlying structure is a graph\cite{FWu}. 
In recent years,  GCNs have been successfully applied to various tasks in data mining, including node classification~\cite{Kipf}, recommendation systems~\cite{YingKdd}, the prediction of the side effects of combinations of drugs (polypharmacy side effects) ~\cite{zitnik2018modeling}, and the prediction of interfaces between proteins.

In node classification problem, GCNs take as input i) an undirected graph that represents the relationships between the data items (vertices), ii) a feature vector associated with each vertex, and iii) the labels associated with some of the vertices. The objective is to predict the labels of the other vertices in the graph, using the feature vectors (as in standard supervised learning) and based on the premise that the labels are assortative in the network (i.e., the vertices that are in close proximity in the network are likely to be labeled similarly). At each layer, the convolution is performed by applying a first-order spectral filter to the feature matrix, followed by a nonlinear activation function~\cite{FWu}. The spectral filter’s connectivity is based on the connectivity of the graph; thus in effect, the features are smoothed across the graph at each layer of the neural network.

Despite the successful application of GCNs to many important problems, one subtle issue remains unresolved and poses challenges to broader application of GCNs: As in many other deep learning applications, the success of the GCNs relies on the existence of many labeled samples (nodes in our case) ~\cite{AAAI18Li}. 
In many real-world applications, however, only a small fraction of the nodes are labeled. 


Training machine learning algorithms with limited labeled data is a long standing and well-studied problem~\cite{lake2015human}.
However, the problem has attracted less attention in the context of GCNs.
To the best of our knowledge, there is only one study that aims to address
this issue for GCNs~\cite{AAAI18Li}. 
Observing that GCNs rely on the assortativity of labels, Li {\em et al.} propagate
labels across the network {\em before} training the GCN, thus increasing the number of labeled samples. 
They use a random walk based approach,  ParWalk~\cite{WuX}, to identify unlabeled nodes that are in close proximity to the labeled nodes, and label these new nodes to expand the set of labeled nodes. 
While this approach has been shown to be effective in enabling the application of GCNs to instances with few training instances, it overlooks an important problem that is associated with the application of random walk based techniques: Random walk based assessment of network proximity assigns higher
scores to nodes with high connectivity and/or centrality~\cite{dada, CoskunLink}, thus biasing the set of labeled nodes toward highly connected nodes.

Building on our earlier work in the context of link prediction ~\cite{CoskunLink}, here we propose a label expansion algorithm, \algo, that aims to fairly assess the similarity of the nodes in a graph, i.e., without being influenced by individual factors such as the connectivity of individual nodes. \algo\ is based on the premise that nodes that are topologically ``similar” are likely to be similar to each other in terms of their proximity to other nodes in the graph. In other words, as opposed to directly assessing the proximity of two nodes in the graph, we assess their similarity in terms of what they are close to. As we have shown previously in the context of prioritizing candidate disease genes~\cite{vavien} and link prediction~\cite{CoskunLink}, this approach drastically reduces degree bias in assessing the topological similarity between the nodes
of a graph. 

While it is useful to assess topological similarity by comparing proximity profiles, the proximity profiles become very high-dimensional
for very large graphs.
To address this problem, instead of assessing the proximity of each node
to every other node in the graph, we assess the proximity of each node to the ``communities" which are potential representatives of graph topology.
For this purpose, we first use a graph clustering algorithm to identify communities in the graph.
Subsequently, we compute reduced topological profiles for each node based on its proximity to the communities. 
Comparing the reduced topological profiles of candidate nodes to that of
the labeled nodes, we identify the nodes that are most topologically similar to the labeled nodes, and label those nodes accordingly.
Since the graphs we consider are very large, computing topological profiles for all nodes in the graph and comparing these profiles to that of the label nodes can be computationally costly. 
To address this efficiency problem, we develop two heuristic algorithms \tptrain\ and \mltrain\ 
that leverage (i) random walk based proximity and (ii) manifold learning, respectively. 
The core idea behind the first heuristic, \tptrain,  is to use random walk based proximity to identify candidate nodes and consider only the nodes in that limited set as potential nodes for labeling. 
The second heuristic, \mltrain, idea relies on finding fractionally added nodes by manifold learning independent from the random walk, then identifying the nodes that are topologically similar to the labeled nodes.

To test the performance of the proposed algorithms in improving the accuracy of GCN-based node classification, we perform comprehensive computational experiments on three citation networks that are often used for benchmarking GCNs. 
Our results show that the proposed label expansion algorithms render
GCNs highly effective in node classification, and the resulting algorithms outperform Random-Walk based label expansion methods~\cite{AAAI18Li}.

The rest of the paper is organized as follows: In Section~\ref{sec:method}, we describe the terminology, establish background on GCNs and label expansion approaches for training GCNs, and describe our method.In Section~\ref{sec:Expreiments}, we provide detailed experimental evaluation of our methods. We draw conclusions and summarize avenues for further research in Section~\ref{sec:conclusion}.
\section{Methods}
\label{sec:method}
\begin{figure*}[t]
	\centering
	\scalebox{0.5}{\input{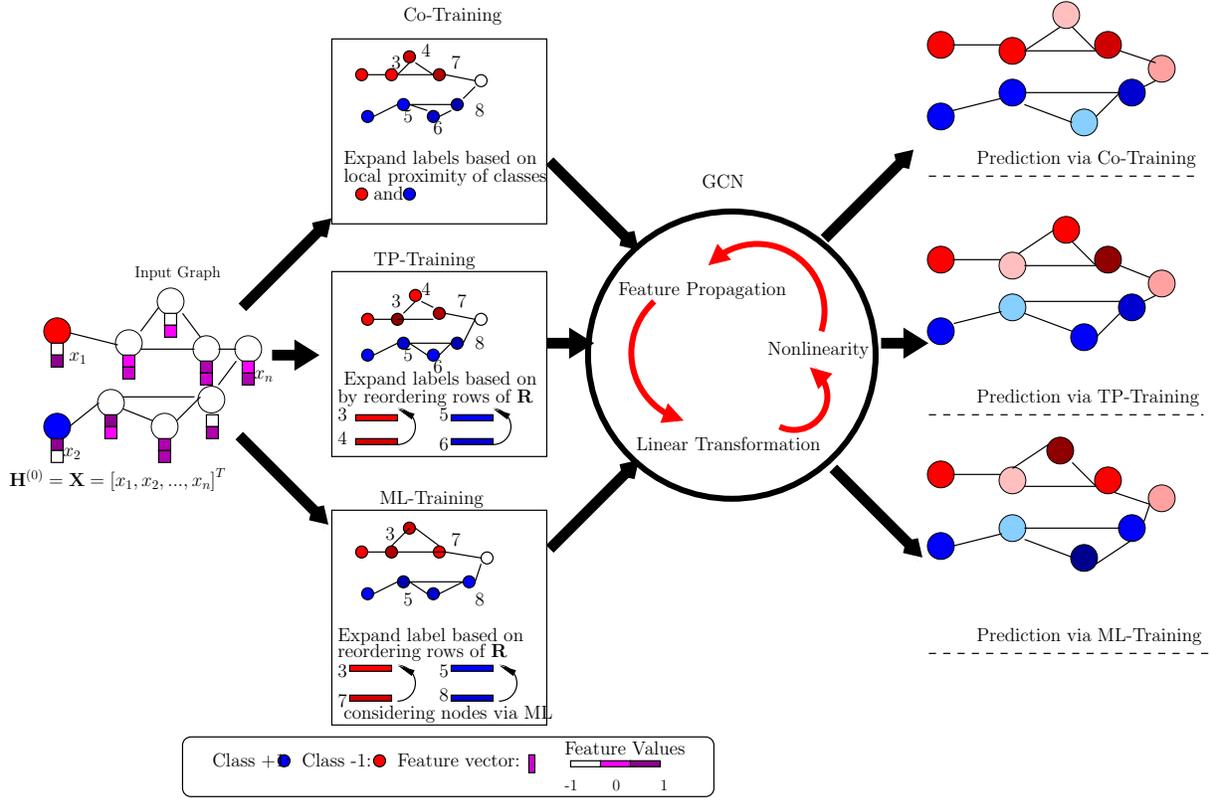}}
	\caption{\textbf{Flowchart illustrating the proposed algorithm, \algo, for expanding set of labeled nodes to train a Graph Convolutional Network (GCN).} Given an undirected graphs with features associated with all nodes and labels associated with some nodes (red and blue nodes in the first graph), \algo\ performs the following steps: (i) clustering of the network to identify communities, (ii) use of network diffusion algorithms to obtain a low-dimensional topological profile for each node, (iii) comparing these topological profiles to identify nodes that are most similar to the labeled nodes. Here, \tptrain\ uses random walk based proximity for expansion, whereas \mltrain\ uses manifold learning for the same purpose. Among the candidate nodes, those that are topologically most similar to the labeled nodes are identified and labeled accordingly. Finally, the GCN is trained using the expanded set of labeled nodes.
} 
\label{fig:mainFigure}
\end{figure*}

In this section, we first define the node classification problem and graph convolutional networks(GCNs). We then present insights for the usage of global topology of the underlying graph to exploit the assortativity of node labels, with a view to exanding the set of labeled samples. Subsequently, we show that considering global correlation  along with local proximity with respect to the labeled nodes can be integrated into the extrapolation of node labels. 
Finally, we propose two heuristic algorithms to implement this approach. 

\subsection{Preliminaries.}
Here, we follow the setting introduced by Kipf and Welling~\cite{Kipf} to present GCNs in the context of the node classification problem. 
In a nutshell, GCNs take as input an undirected graph with labels on some of the nodes, as well as feature vectors associated with each node, and output the labels for all nodes in the graph based on two premises: (i) (all or some of) the features are correlated with labels (as in any supervised machine learning setting), (ii) labels are assortatively distributed with respect to graph topology (i.e., nodes that are neighbors are likely to have the same label)~\cite{Kipf}.

\textbf{Problem (Node Classification):} We are given an undirected graph ${\network} ={({\nodeset},{\edgeset}, {\XMatrix})}$, where $\nodeset$ denotes the set of nodes and $\edgeset$ denotes the set of edges. We are also given a feature matrix ${\XMatrix} = {[x_1, x_2, ... ,x_n]} \in \mathbb{R}^{n \times d}$ such that $x_i \in \mathbb{R}^{d}$ is a feature vector for node $v_i \in {\nodeset}$, where $n=|\nodeset|$ denotes the number of nodes. For a given set of labeled nodes, $\mathcal{V}_{l} \subset \nodeset$, our objective is to assign labels to the remaining nodes, $\mathcal{V}_{u} = \mathcal{V} - \mathcal{V}_{l}$. The set of classes is denoted $C$, where $k=|C|$ is the number of classess. The class assignments of labeled nodes are given by a $|\mathcal{V}_l| \times k$ binary matrix $\mathbf{Y}$, where $y_{ij} = 1$ indicates that
node $v_i \in \mathcal{V}_{l}$ belongs to class $j \in C$.


\subsubsection{Graph Convolutional Networks.} 
GCNs are simplified models of graph convolutional neural networks (GCNNs), which are generalizations of conventional convolutional neural networks (CNNs) on graphs ~\cite{AAAI18Li}. 
With a logic similar to that of CNNs, given the feature vectors of all nodes in the graph, GCNs learn a new feature representation for each node in the graph  over multiple neural network layers which are then used as input to final classifier~\cite{FWu}. 
In the GCN, the input to the $\ell$th graph convolution layer is an activation matrix denoted
$\HMatrix^{(\ell-1)}$ and the output of the layer is the activation matrix denoted $\HMatrix^{(\ell)}$~\cite{FWu}. The input to the initial layer is therefore the feature matrix, i.e.,:
\begin{equation}
\HMatrix^{(0)} = \XMatrix
\label{eq:HeqX}    
\end{equation}
In each graph convolution layer, $\HMatrix$ is updated in three steps: feature propagation, linear transformation, and the application of a nonlinear activation function~\cite{FWu}.

\textbf{Feature Propagation.} is the process of propagating the across the graph. 
More specifically, in each layer, the incoming features of each node 
$v_i \in {\nodeset}$ are aggregated  with the incoming features of the nodes 
that are in the vicinity of $v_i$ in ${\network}$~\cite{FWu}.
Using the notation of Li {\em et al.}~\cite{AAAI18Li}, we can express this update over the entire graph as a matrix operation. 
Namely, defining the convolution matrix as
$\hat{\AMatrix} = {\Tilde{\DMatrix}^{-1/2}} \Tilde{{\AMatrix}} \Tilde{{\DMatrix}}^{-1/2}$ where $\Tilde{\AMatrix} = \IMatrix + \AMatrix$ and $\Tilde{\DMatrix}={\IMatrix}+{\DMatrix}$ (i.e., adding self-loops to each node in the adjacency matrix and the diagonal degree matrix)~\cite{AAAI18Li},
the update for all nodes becomes a single matrix multiplication: 
\begin{equation}
\hat{\HMatrix}^{(\ell)} = \hat{\AMatrix}\HMatrix^{(\ell-1)}
\label{eq:HeqAH}    
\end{equation}

Clearly, this step encourages incident nodes to have similar features, which is further
used to make similar predictions for neighboring nodes. 
Li {\em et al.}~\cite{AAAI18Li} show that this step is equivalent to Laplacian smoothing and using many layers in GCNs causes degradation in prediction accuracy because of over-smoothing and the mixing of labels.

\textbf{Linear Transformation and Point-wise Non-linear Activation.}
In each layer of the GCN, once the feature matrix is smoothed across the graph, the resulting intermediary feature
matrix is subjected to linear transformation using a trainable weight matrix $\Theta^{(l)}$. 
Subsequently, a nonlinear activation function, such as $ReLU = max(x,.)$, is used~\cite{FWu} to produce the output activation matrix for that layer:

\begin{equation}
\HMatrix^{(l+1)} = ReLU\big(\hat{\HMatrix}^{(\ell)}\Theta^{(\ell)}\big) 
\label{eq: GCN}    
\end{equation}

\textbf{Node Classification.} The final GCN layer is reserved for predicting the unknown labels of nodes by using a \textit{softmax} classifier. Formally, let  $\hat{\YMatrix} \in \mathbb{R}^{n \times k}$ denote the class prediction matrix, where  $\hat{y}_{ij}$ shows the probability that node $v_i \in {\nodeset}$ belongs to class $j$ for $1 \leq j\leq k$, where $k$ denotes the number of classes. In the final layer, the class prediction matrix is computed as~\cite{FWu}:
\begin{equation}
\hat{\YMatrix} = softmax\big( \hat{\AMatrix} \HMatrix^{(L-1)} \Theta^{(L)}\big)
\label{eq:GCNClassification}    
\end{equation}
where 
$softmax(x) = \dfrac{1}{\sum_{j=1}^{k} exp(x_c)}exp(x) $ transforms
predicted values into a well-defined probability density function~\cite{FWu}.

In our experiments, we use a two-layered GCN as introduced by Kipf and Welling~\cite{Kipf}. The motivation for limiting the network to two layers is to overcome the over-smoothing problem reported by Li {\em et. al}~\cite{AAAI18Li}. The two-layered GCN can be defined in compact form as~\cite{Kipf}: 
\begin{equation}
\hat{\YMatrix} = softmax(\hat{\AMatrix}ReLU(\hat{\AMatrix}\XMatrix\Theta^{(0)})\Theta^{(1)}),
\label{eq:TwoGCN}    
\end{equation}

It is important to note that the proposed approach does not depend on the architecture of the GCN and can be directly applied to other architectures as well.

\textbf{Label Expansion Problem.} 
Despite the demonstrated effectiveness of GCNs in node classification, it has been observed~\cite{AAAI18Li} that GCNs require a large number of labeled nodes to train the model. 
Li {et al.}~\cite{AAAI18Li} analytically characterize the number of nodes
required to effectively train a $\tau$-layered GCN as follows:
Let $t$ denote the number of labeled nodes that are associated with a given class.
If the average node degree in the graph is $\bar{\delta}$, we must have $\bar{\delta}^{\tau} \times t \geq n$ for the GCN to effectively propagate features across the entire graph. 
Solving for $t$, we obtain $t^{*}= \log n/\tau \log \bar{\delta}$ as the minimum number of nodes required to be labeled by a given class in the training data. 

Let $t_j=|\{v_i \in {\mathcal{V}}_l:y_{ij}=1\}$ denote the number of
nodes labeled with class $j \in C$.
If $t_j < t$, it is necessary to expand the set of labeled nodes for class $j$ using a method that does not require training. 
Motivated by this observation, the {\bf Label Expansion Problem} 
is defined as  the problem of finding $t-t_j$ additional nodes to be labeled by class $j$ to facilitate the training of the GCN.

\subsubsection{Existing Solution to the Label Expansion Problem.}
Li {\em et al.}~\cite{AAAI18Li} propose a random walk based approach, \textbf{\cotrain}, to expand the set of labeled nodes {\em before} training the GCN.
They show that this approach clearly outperforms the alternate approach of 
iteratively expanding the set of labeled nodes by repeatedly training GCNs and
using the predictions to expand the set of labeled nodes.
To expand the set of labeled nodes before training, they use partially absorbing random walks (ParWalks), which is a second-order Markov chain with partial observation at each node~\cite{WuX}. It has been shown that ParWalk~\cite{WuX} can capture the global structure of graph and gives better node raking results than classical random walks, such as PageRank~\cite{WuX}. 
As with many other random-walk based algorithms, ParWalk has a closed-form solution and can be formulated as a linear system of equations as follows:
\begin{equation}
\PMatrix = \LMatrix + \alpha\Lambda.
\label{eq:coeff}    
\end{equation}
Here, ${\LMatrix} = {\DMatrix} - {\AMatrix}$, denotes the Laplacian and $\alpha$ (scalar) and $\Lambda $
($n \times n$ matrix) are parameters to be tuned (which are chosen as respectively $ 10^{-6}$ and $\IMatrix$ by Wu {et al.}~\cite{WuX}).

Given the set of labeled nodes $\mathcal{V}_{l}$, Li {\em et al.}~\cite{AAAI18Li} use the inverse of $\PMatrix$ to identify new nodes 
to be labeled. Namely, for each class $j$, they compute an $n$-dimensional vector as follows: 
\begin{equation}
\textbf{p}_j = {\sum_{v_i:y_{ij}=1}} \PMatrix^{-1}_{:,j},
\label{eq:pvec}    
\end{equation}
thus $\textbf{p}_j (i)$ indicates the proximity of node $v_i \in \nodeset$
to the nodes in class $j$.
Consequently, for each class $j \in C$, \textbf{\cotrain} labels the $t-t_j$ nodes with largest values in  $\textbf{p}_j (i)$. 
Then, the GCN is trained with newly added labels~\cite{AAAI18Li}.

\subsection{Proposed Solution to the Label Expansion Problem.} 
Here, we stipulate that the assortativity (of labels) in a graph can be
exploited more effectively by quantifying the {\em relative} position of
nodes in the graph with respect to each other. 
In other words, instead of asking the question ``is a node close to the labeled nodes in the network?", we ask the question ``does the node see the network from a perspective similar to that of the labeled nodes?"
In the context of link prediction~\cite{CoskunLink} and its applications to various problems in computational biology (e.g., candidate disease gene prioritization~\cite{vavien}, drug response prediction~\cite{CoskunDrug}), we have shown that this approach is indeed more effective than direct consideration of proximity, in that it drastically reduces bias caused by individual-node related factors, such as connectivity and/or centrality~\cite{vavien}.
Elimination of such bias is particularly important for the label expansion problem, since expansion of labels toward a biased set of nodes would misguide the entire training process.

The assessment of the relative positions of nodes with respect to each other in a graph requires computation of ``topological profiles" for each node in the graph, followed by the comparison of these profiles to assess relative positions. 
For label expansion in GCNs, the graphs considered are rather large, thereby posing challenges associated with high-dimensionality and computational complexity.
To tackle these challenges, we here propose a method that uses {\em graph communities} as {\em landmarks} to compute topological profiles for each node.
Namely, the proposed method, named {\bf Label Expansion Using Community Landmarks} (\algo), first identifies communities in the graph, subsequently computes topological profiles for each node using their proximity to the communities, and finally assesses the {\em topological similarity} of each node to the nodes that are already labeled. 
It then expands the labels, for each class, by selecting the nodes that are most topologically similar to the nodes that are already labeled with that class. Figure~\ref{fig:mainFigure} depicts the ideas proposed in the algorithm.

There exist many algorithms for identifying communities in graphs~\cite{catalyurek1999hypergraph}. 
Let $K$ be a parameter to be tuned, denoting the number of communities.
\algo\ first identifies communities using an existing graph clustering algorithm,  {\sc GMine}~\cite{rodrigues2006gmine} 
(this can be replaced by any graph clustering algorithm).
$S_1, S_2, ..., S_K$ in $\network$, such that $\bigcup_{1 \leq i \leq K}{S_i}=\nodeset$ and $S_i \cup S_j = \emptyset$ for each $1 \leq i,j \leq K$,
and the nodes within each community (or cluster) are as tightly connected with each other as possible.
For each cluster $S_i$, we compute an $n$-dimensional  vector ${\mathbf{r}_{S_i}}$, representing the proximity of each node to the nodes in
$S_i$ as follows: 
\begin{equation}
{\mathbf r}_{S_i} = \IMatrix + \PMatrix\times e_i + \PMatrix^2\times e_i +...+ \PMatrix^m\times e_i
\label{eq:ParWalkForClusters}    
\end{equation}
Here $e_i$ is the vector that attracts the random walk toward the nodes in $S_i$, defined as $e_i(j)=1/|S_i|$ if $v_j \in S_i$ and $e_i(j)=0$ otherwise.
Since $\PMatrix$ (defined in Equation (\ref{eq:coeff})) is a symmetric positive definite matrix~\cite{WuX}, we can approximate its inverse through addition of $m$ iterative multiplications, where $m$ denotes the dimension of the Krylov subspace~\cite{CoskunVldb}. 

Computing ${\mathbf{r}_{S_i}}$ for each cluster $1 \leq i \leq K$,  we obtain {\em topological profile matrix} $\RMatrix \in \mathbb{R}^{K\times n}$ which contains $\mathbf{r}_{S_i}$ in its $i$-th column.
The matrix $\RMatrix$ provides a reduced-dimensional representation of the
global topology of the graph, in terms of proximities of nodes to the communities in the graph. 
\begin{algorithm}
\caption{Expand the Label Set via \algo\ }
\label{alg:PCAlgo}
\begin{algorithmic}[1]
\State Partition $\network$ into $K$ clusters~\cite{rodrigues2006gmine} \Comment{Offline}
\State Define $\PMatrix = \LMatrix + \alpha \Lambda$ \Comment{Offline}
\State Construct $\RMatrix\in \mathbb{R}^{K\times n}$ matrix \Comment{Offline}
\For{\texttt{each class $j$}}
\For{\texttt{$v_i \in {\nodeset}_{u}$}}
      \State \texttt{${\mathbf{b}}_j(i) = \sum_{v_{\ell}:  y_{\ell j}=1} \rho(\RMatrix(:,\ell), \RMatrix(:,i))$}
      \EndFor
      \State Find top $t-t_j$ nodes in ${\mathbf{b}}_j$
       \State Add them to training set with label $j$
 \EndFor

\end{algorithmic}
\end{algorithm}

\subsubsection{Baseline Algorithm for Topological-Profile Based Label Expansion.}
Once the topological profile matrix $\RMatrix$ is computed, the nodes in the graph that are topologically ``similar" to the labeled nodes can be identified by comparing the respective columns of $\RMatrix$.
This approach is shown in Algorithm~\ref{alg:PCAlgo}. 
After pre-processing (graph clustering and computation of $\RMatrix$), for each class $j \in C$, we compute a topological similarity score for each node 
$v_i \in {\nodeset}_{u}$, indicating the similarity of $v_i$'s topological profile to that of the nodes that are labeled $j$:

\begin{equation}
    {\mathbf{b}}_j(i) = \sum_{v_{\ell}:  y_{\ell j}=1} \rho(\RMatrix(:,\ell), \RMatrix(:,i))
\end{equation}
Here, $\rho(.,.)$ denotes Pearson's correlation.
Finally, we identify the top $t-t_j$ entries in ${\mathbf{b}}_j(i)$, 
and label the respective nodes with $j$.

Although Algorithm~\ref{alg:PCAlgo} demonstrates the use of topological similarity in identifying nodes for label expansion, one subtle issue is that lines $4-6$ of the algorithm require exhaustive computation of $|{\nodeset}_l \times n$ correlations of
$k$-dimensional vectors. 
In our experiments, we observe that this computation dominates the computation costs of GCN training, resulting in a more effective but slower algorithm than \textbf{\cotrain}~\cite{AAAI18Li}. 
To remedy this efficiency problem, we propose two heuristics that implement the same
idea with optimizations that drastically improve runtime without compromising effectiveness.

\subsubsection{Heuristic Algorithms for Topological Profile Based Label Expansion.}
The first heuristic we propose, \tptrain, reduces the number of correlation computations
by focusing on a smaller candidate set of nodes for expansion.
Observing that a node that is topologically similar to the labeled nodes (i.e., have a high value in $\mathbf{b}_j$ is also likely to be proximate to these nodes (i.e.,  have a high value in $\mathbf{p}_j$), we use the proximity vector computed by \cotrain\ to select candidate nodes. 
In \tptrain, for each classs $j \in C$, we first compute $\mathbf{p}_j$ 
(unlike \cotrain, we use the conjugate gradient algorithm~\cite{demmel1997applied} to solve the linear system of equations).
Subsequently, letting $\eta$ denote the {\em fraction of additional candidate nodes} considered, we identify the top $(1+\eta)t$ nodes with
largest entries in $\mathbf{p}_j$.
We then compute the topological similarity of these $(1+\eta)t$ nodes to the nodes labeled by $j$.
Among these, we identify the $t$ nodes with highest topological similarity and label these nodes by $j$.
This algorithm is shown in Algorithm~\ref{alg:HeuristicAlgo}.

\begin{algorithm}
\caption{\tptrain }\label{alg:HeuristicAlgo}
\begin{algorithmic}[1]
\State Partition $\network$ into $K$ clusters~\cite{rodrigues2006gmine} \Comment{Offline}
\State Define $\PMatrix = \LMatrix + \alpha \Lambda$ \Comment{Offline}
\State Construct $\RMatrix\in \mathbb{R}^{K\times n}$ matrix \Comment{Offline}

 \For{\texttt{each class $j \in C$}}
      \State \texttt{$\mathbf{p}_j = {\sum_{v_i:y_{ij}=1}} \PMatrix^{-1}_{:,j}$}
      \State Find the top $(1+\eta)t$ nodes in $\mathbf{p}$
       \For{\texttt{each node $v_i$ among these nodes}}
        \State \texttt{${\mathbf{b}}_j(i) = \sum_{v_{\ell}:  y_{\ell j}=1} \rho(\RMatrix(:,\ell), \RMatrix(:,i))$}
         \EndFor
        \State Find top $t-t_j$ nodes in ${\mathbf{b}}_j$
       \State Add them to the training set with label $j$
 \EndFor
\end{algorithmic}
\end{algorithm}

The second heuristic algorithm we propose, \mltrain, also identifies $(1+\eta)t$ nodes to be considered for their topological similarity with the labeled nodes.
The first $t$ of these nodes are the nodes selected by \cotrain\ for expansion, i.e., the nodes that are closest to the nodes that are labeled.
To select the additional $\eta t$ candidates, however, we do not use information on the labeled nodes.
Instead, we use \textit{manifold learning}~\cite{wachinger2015diverse} on $\RMatrix$ to select $\eta t$ nodes that are diverse in terms of their topological profiles .
\begin{algorithm}
\caption{\mltrain (OFFLINE) }\label{alg:HeuristicAlgo2}
\begin{algorithmic}[1]
\State Input: Given $\RMatrix, \eta t$, and a positive integer $m$
\State Output: A set of indices, $S_{I}$,  for $\eta t$ rows of $\RMatrix$.
\State Set $D = \mathbf{1}_n/n$
\State $A_{I} = \emptyset$

 \For{\texttt{ 1 to $\eta t$ }}
      \State Select a random node $v_i \in {\nodeset}_u$
      $\propto D_i$ 
      \State $A_I = A_I \cup \{i\} $
       \State Compute $\Delta_j = \left\lVert {\RMatrix}(:,i) - {\RMatrix}(:,j)\right\rVert$, $\forall j \in {\nodeset}_u - A_I$
       \State Set $\mathcal{N}_i$ $m$ nearest neighbor based on $\Delta_i$
       \State Update $D_j = D_j \times exp(-{\Delta_j}^{2})/2\sigma^2$, $\forall j\in \mathcal{N}_j$
 \EndFor
\end{algorithmic}
\end{algorithm}

The manifold learning based algorithm for candidate selection, \mltrain, is shown in Algorithm~\ref{alg:HeuristicAlgo2}.
Namely, we start by selecting a random node $v_i$, where the selection probability is uniform initially. 
Then, using the topological profiles in $\RMatrix$, we compute the distance of $v_i$ against all other nodes and discard K-nearest neighbors(K-NN), with $K=8$ in our case. 
Finally, we update the selection probabilities based on their geodesic distance with $v_i$ and repeat until  $\eta t$ nodes are selected.
This way, we select rows that are geophysically distant from each other and thus diverse. 
We repeat this process for each class.  Importantly, this row selection via manifold learning can be performed offline since it is independent from labeled nodes.  


\section{Experimental Results}

\label{sec:Expreiments}
\begin{table}[t]
\caption{Descriptive Statics of Datasets}
\label{datatable}
\centering
\resizebox{0.4\textwidth}{!}{%
\begin{tabular}{|c| c c c c|}
\hline
Network &  Nodes &  Edges &  Classes &  Features  \\
\hline\hline
 {\tt Cora} & 2708 &  5429 &  7 & 1433  \\
 {\tt CiteSeer} & 3327  &  4732 & 6  & 3703  \\
 {\tt PubMed} & 19717 & 44338  & 3 & 500  \\
\hline
\end{tabular}
}
\end{table}
\begin{figure*}[htbp]
  \begin{center}
     \subfigure[]{\label{fig:Cora100}
     \includegraphics[scale=0.4]{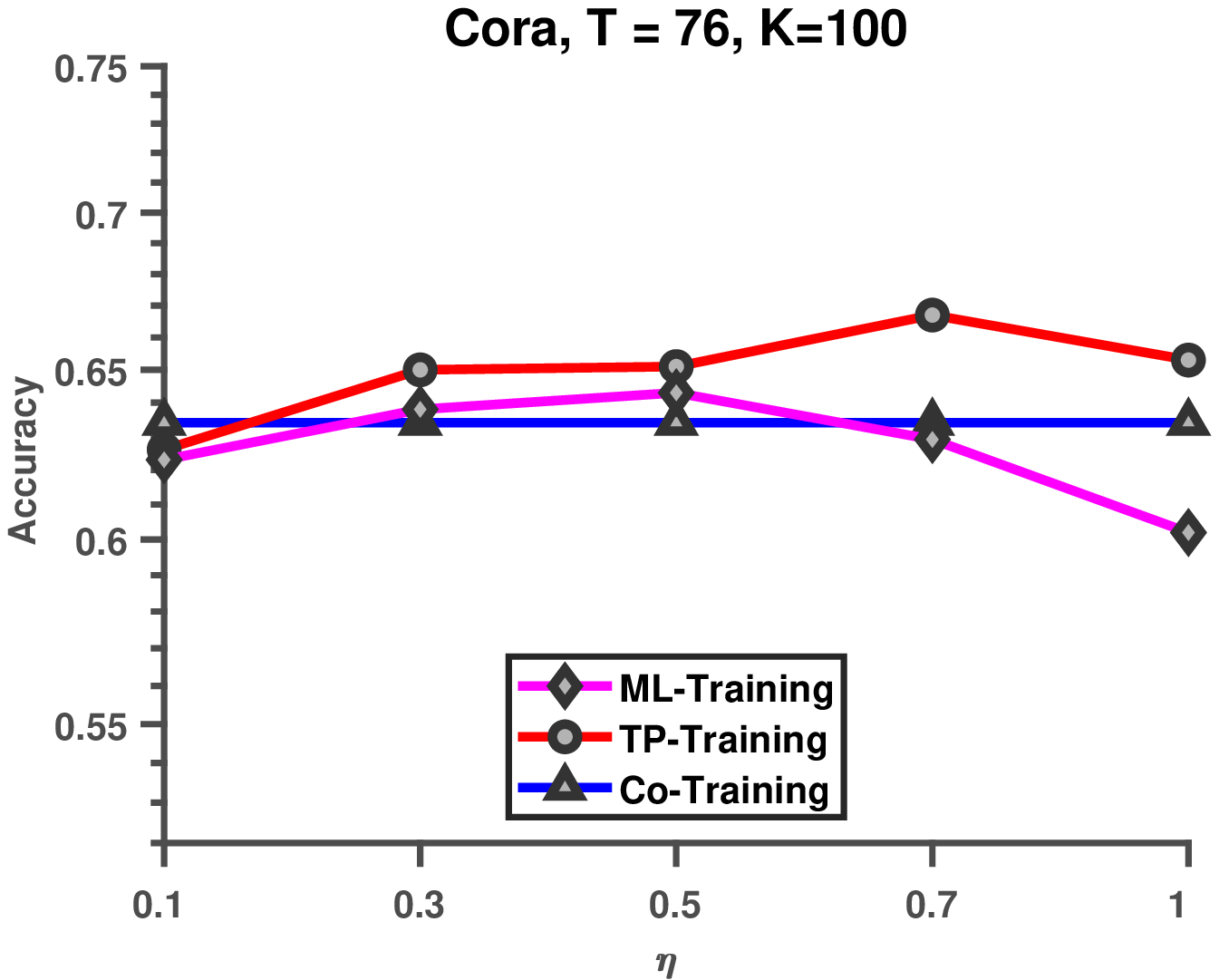}}\quad
     \subfigure[]{\label{fig:Cora250}
     \includegraphics[scale=0.4]{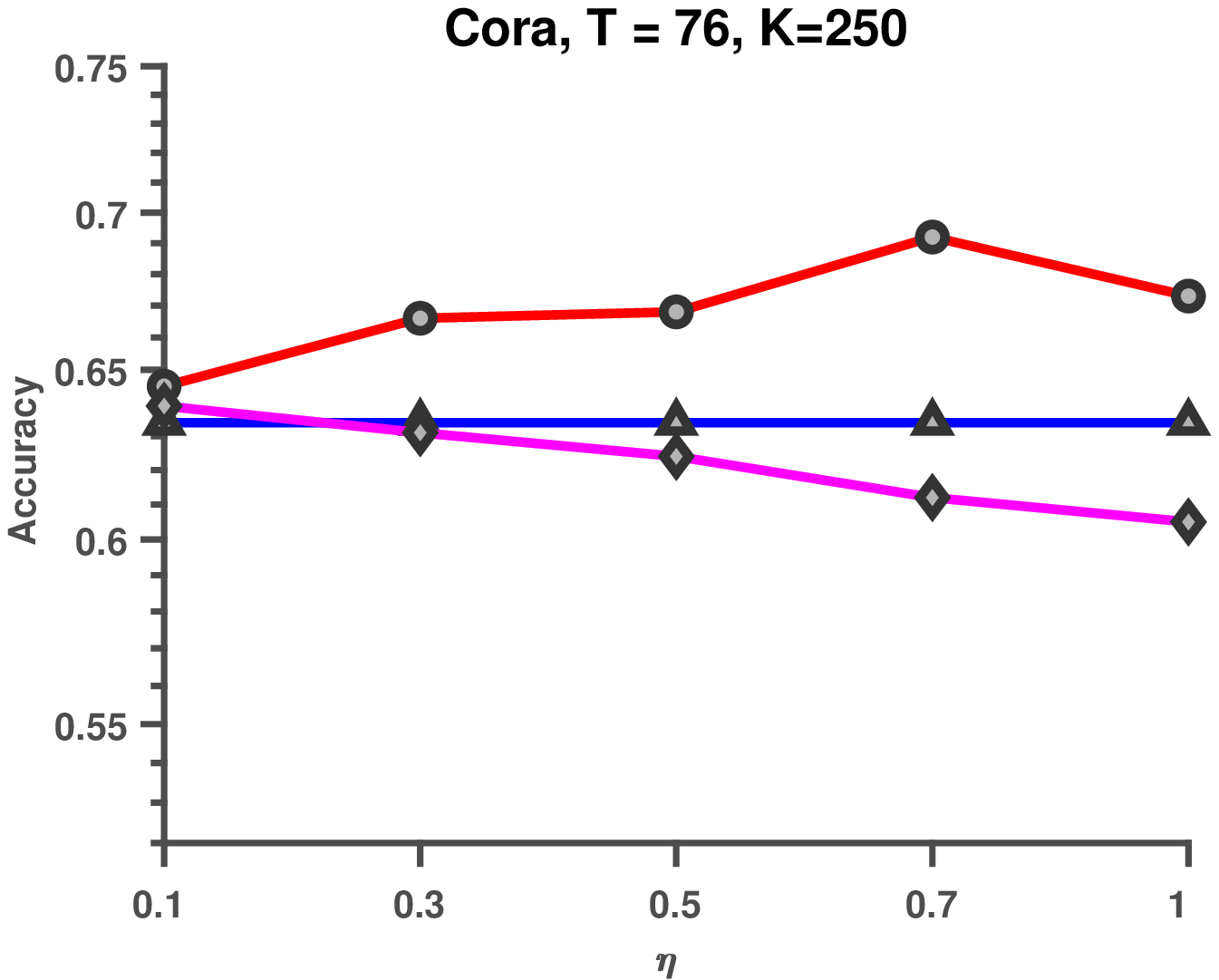}}\quad
      \subfigure[]{\label{fig:Cora500}
      \includegraphics[scale=0.4]{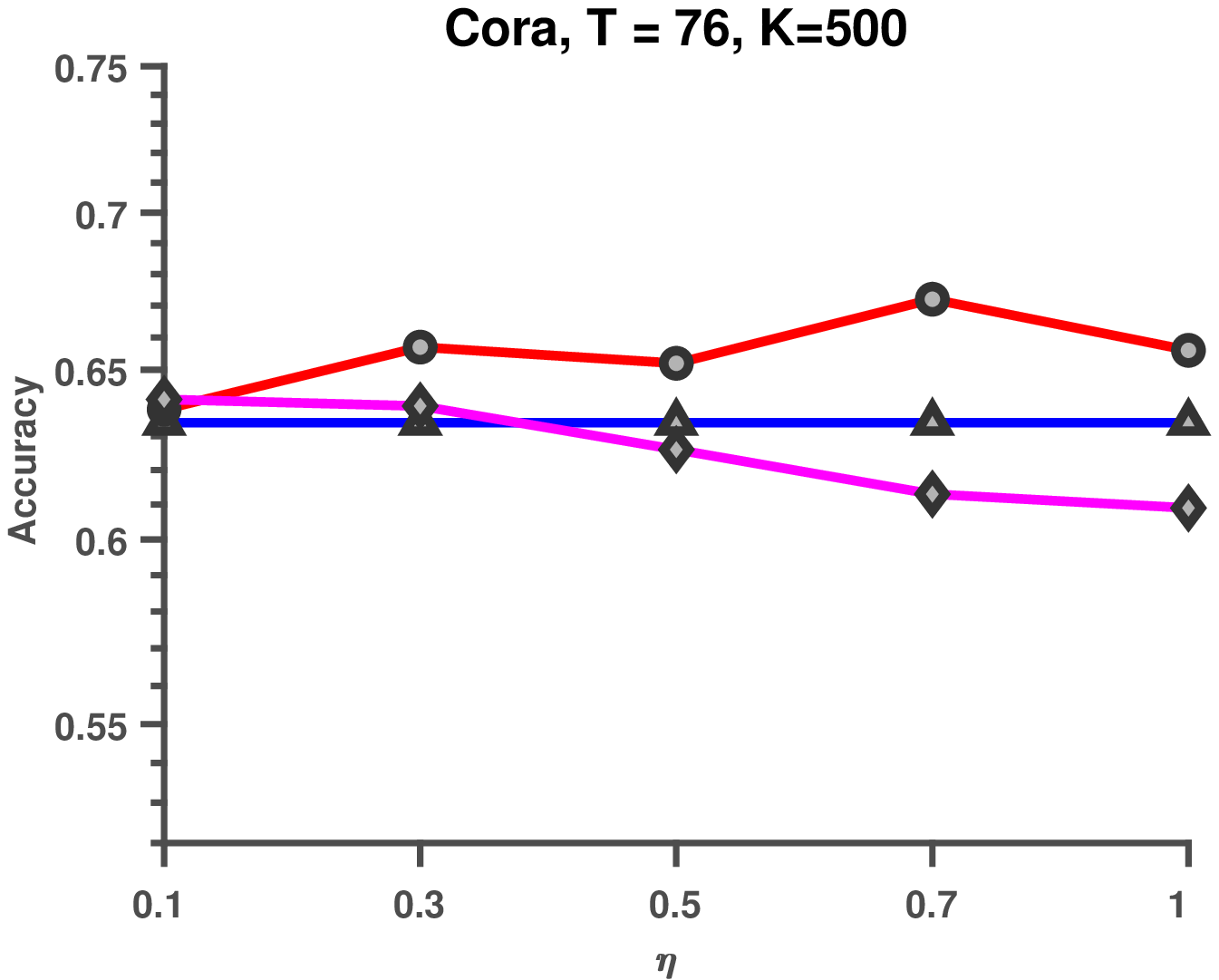}}\quad
    \subfigure[]{\label{fig:Cite100}
     \includegraphics[scale=0.4]{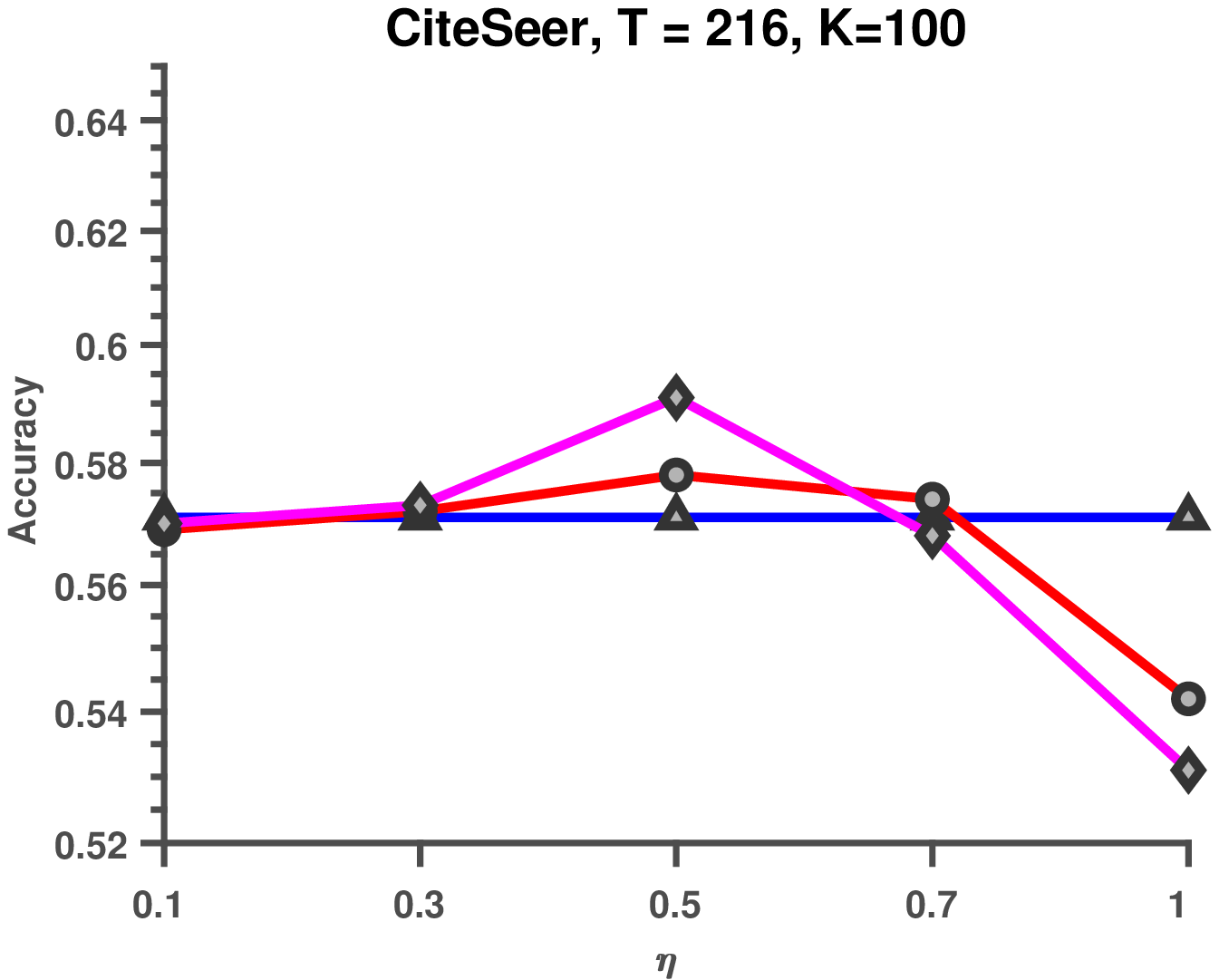}}\quad
     \subfigure[]{\label{fig:Cite250}
     \includegraphics[scale=0.4]{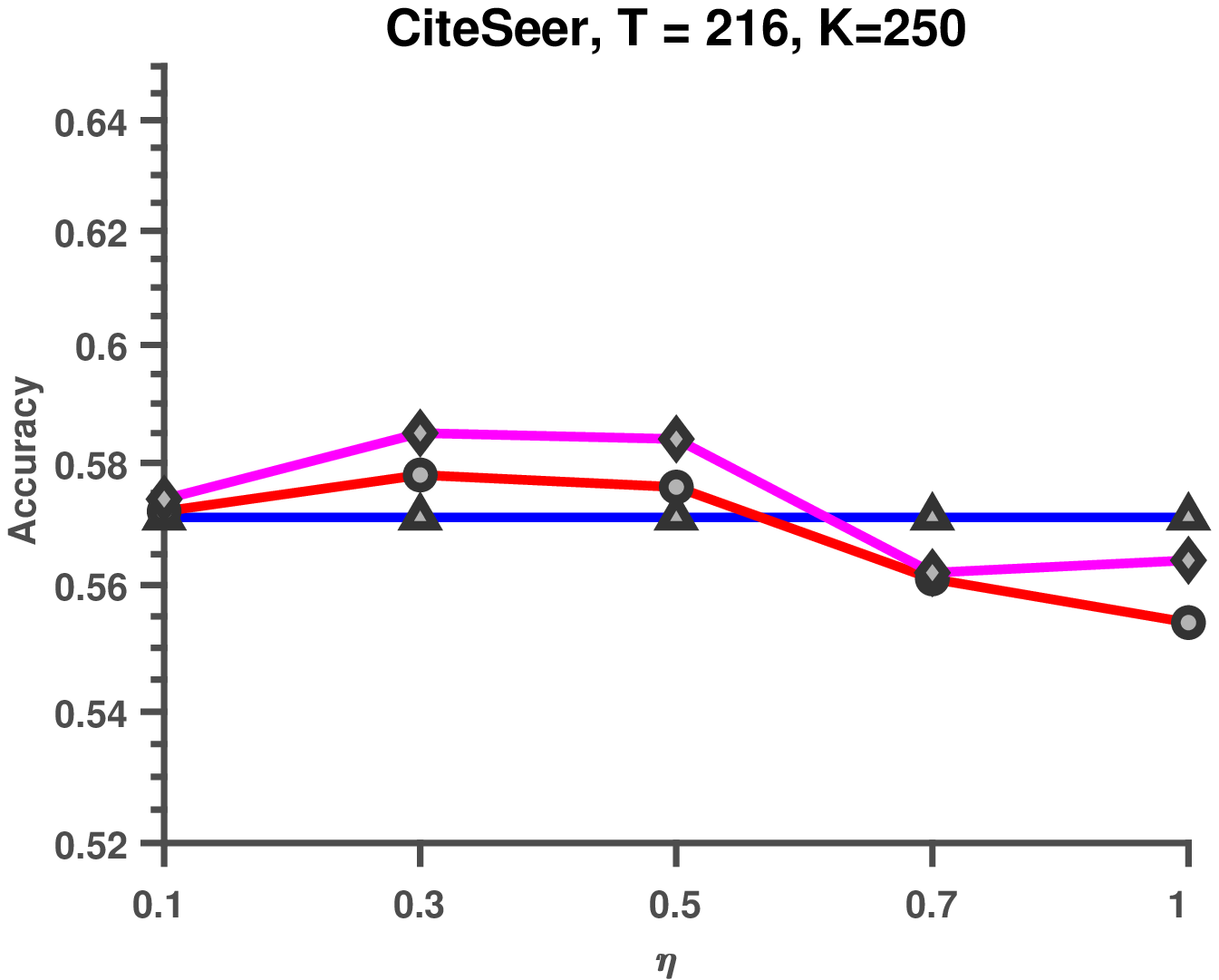}}\quad
      \subfigure[]{\label{fig:Cite500}
      \includegraphics[scale=0.4]{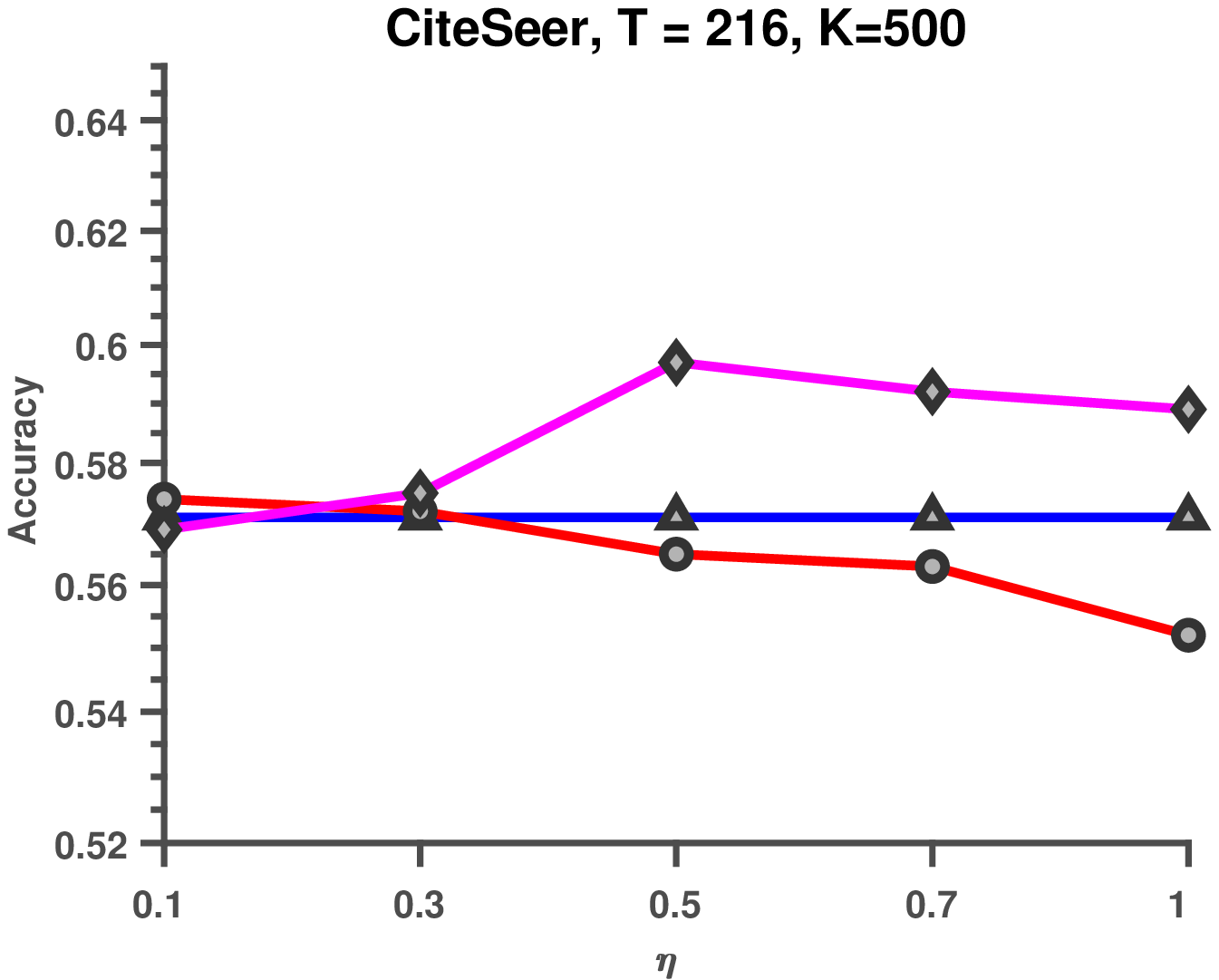}}\quad
    \subfigure[]{\label{fig:Pub100}
     \includegraphics[scale=0.4]{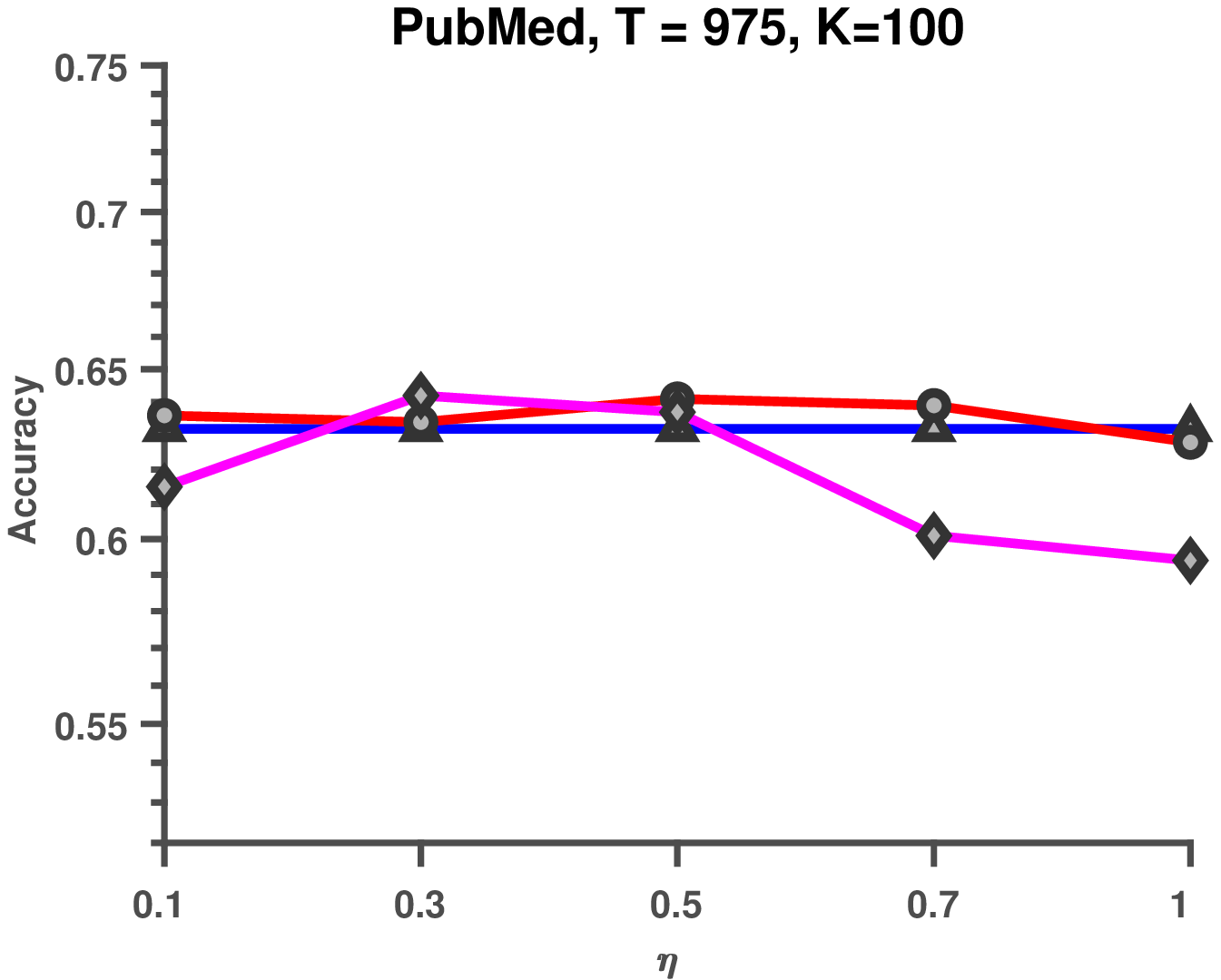}}\quad
     \subfigure[]{\label{fig:Pub250}
     \includegraphics[scale=0.4]{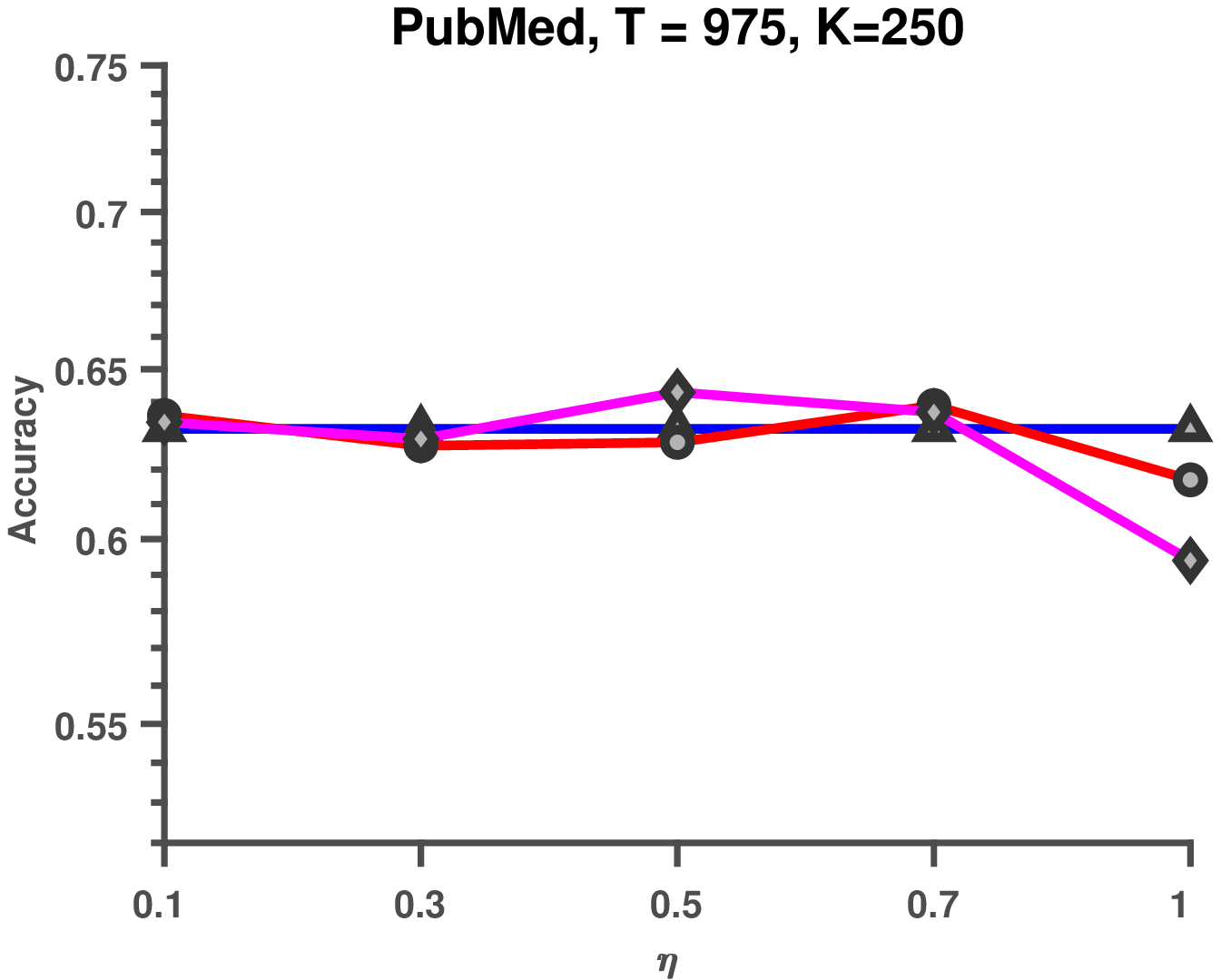}}\quad
      \subfigure[]{\label{fig:Pub500}
      \includegraphics[scale=0.4]{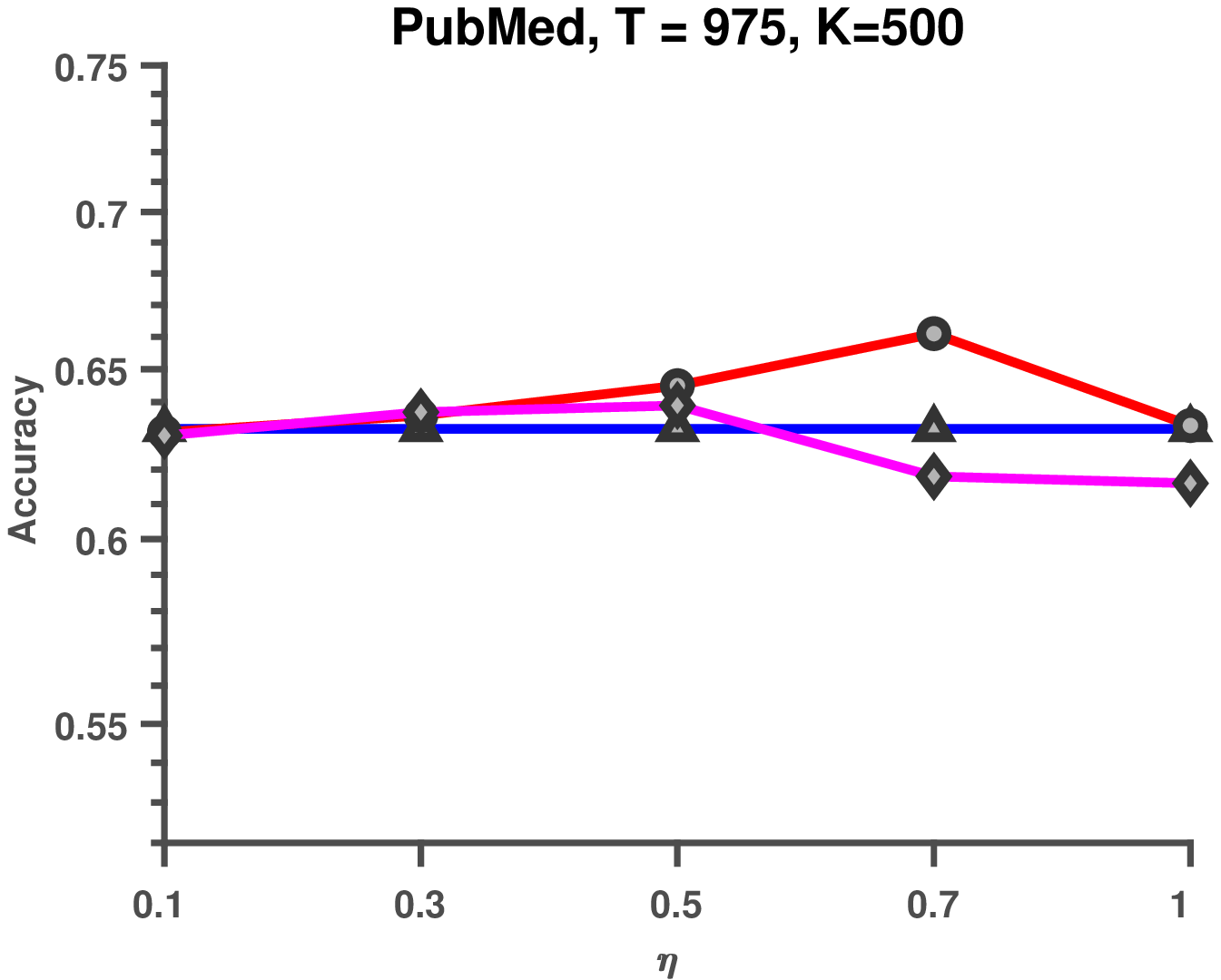}}\quad
   \end{center}
  
  \caption{\textbf{ Comparison of the performance of proposed label expansion algorithms \tptrain\  and \mltrain\ against that of the existing label expansion algorithm \cotrain\ on three  datasets.} The plots show the mean accuracy of a GCN trained using labels expanded by each of the three methods.  The performance of \tptrain\ and \mltrain\ is shown as a function of the fraction of candidates that are considered by label expansion, $\eta$, sampled at $0.1, 0.3. 0.5, 0.7$, and~$1$. Each row corresponds to a dataset and each column corresponds to a different value of the number of communities used to compute topological profiles ($K=100$, $250$, and $500$). \cotrain's performance does not depend on these parameters.}
  \label{fig:Performance}
\end{figure*}
 \begin{figure*}[htbp]
   \begin{center}
     \subfigure[]{\label{fig:Cora}
     \includegraphics[scale=0.4]{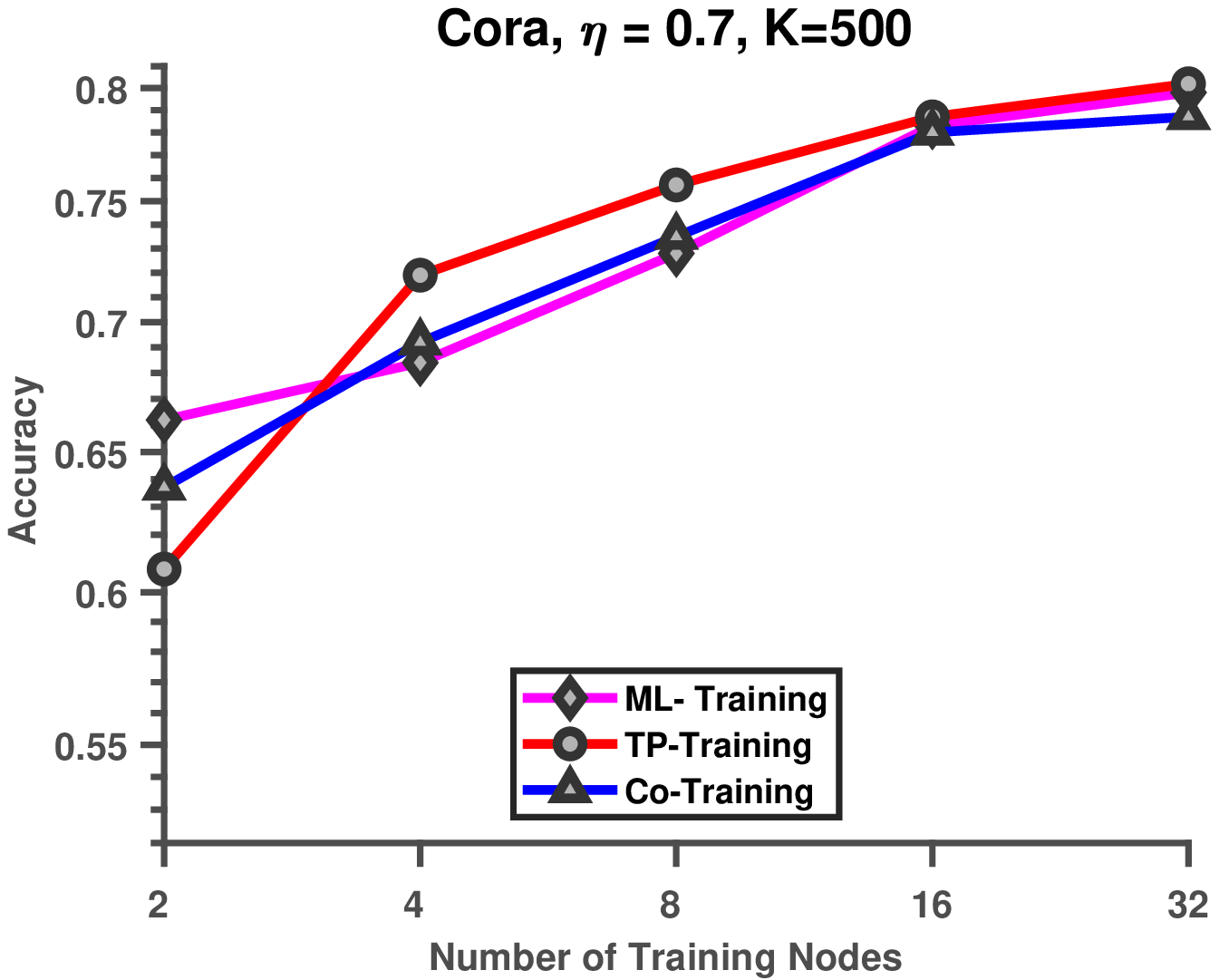}}\quad
     \subfigure[]{\label{fig:CiteSeer}
     \includegraphics[scale=0.4]{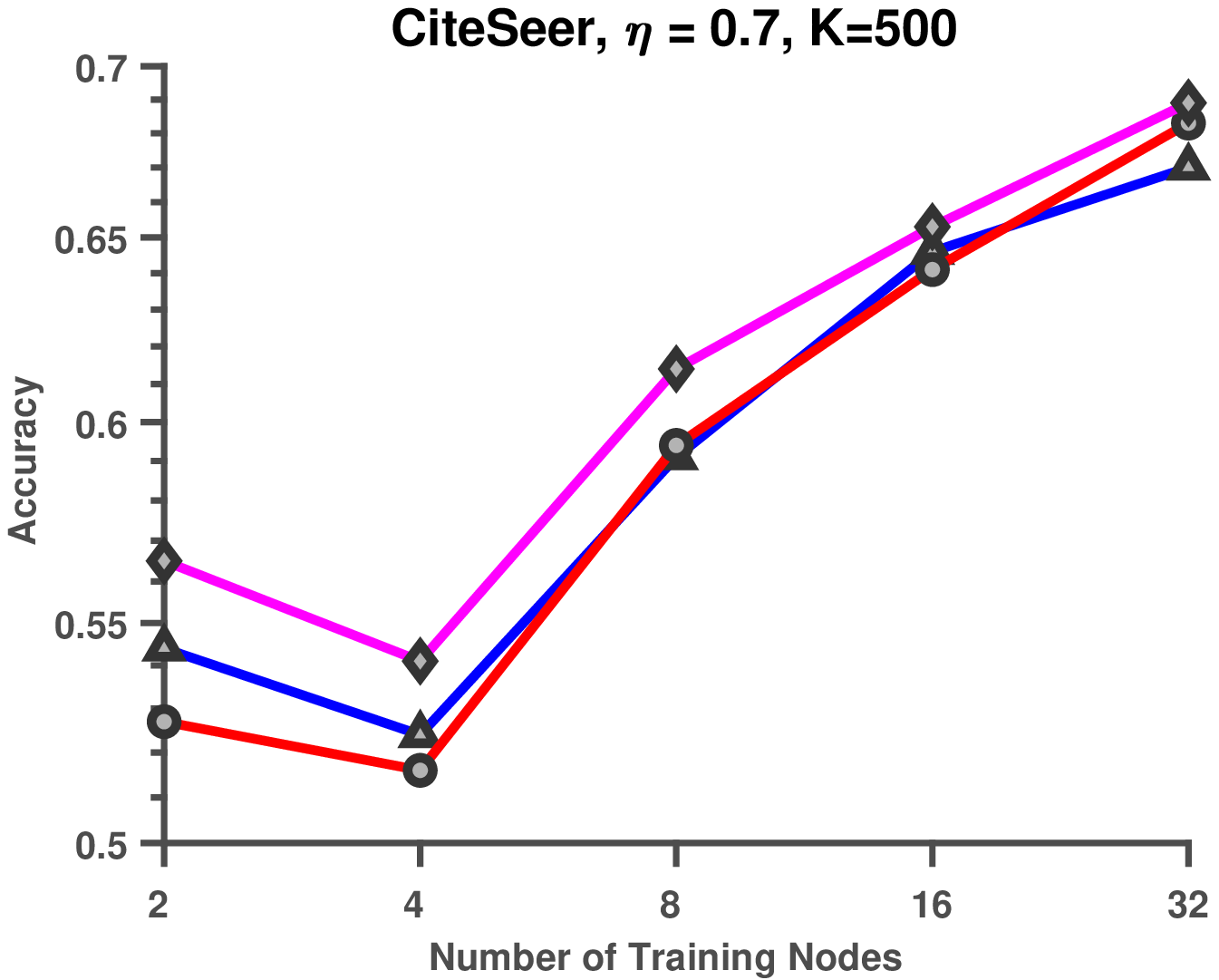}}\quad
      \subfigure[]{\label{fig:PubMed}
      \includegraphics[scale=0.4]{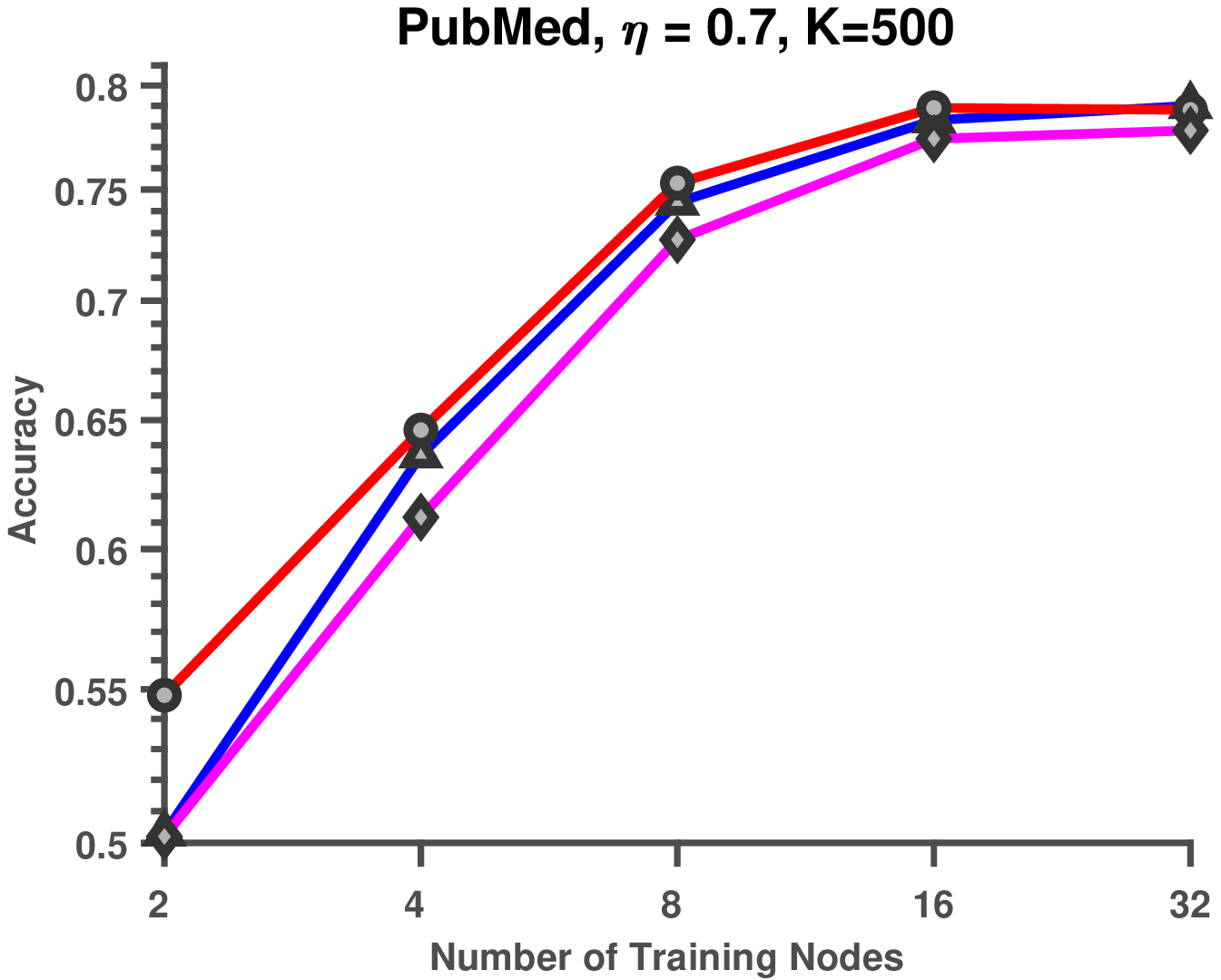}}\quad
   \end{center}
\caption{\textbf{The accuracy of GCNs trained using node labels expanded by proposed algorithms \tptrain\ and \mltrain, as well as existing algorithm \cotrain, as a function of the number of labeled nodes available for each class.} In these experiments, for \tptrain\ and \mltrain, fraction of additional nodes and number of clusters are fixed as $\eta = 0.7$ and $K = 500$ and the reported numbers are the mean of accuracy of GCNs across 10 runs.}
   \label{fig:Performance2}
 \end{figure*}
In this section, we systematically evaluate the performances of proposed \tptrain\ and \mltrain\ algorithms in expanding the set of labeled nodes in the context of the node classification problem. 
We start our discussion by describing the datasets and our experimental setting. 
Next, we analyze the performance of the algorithms as a function of the key parameters. 
We also compare the performance of our proposed methods against the only
other existing method for label expansion, \cotrain~\cite{AAAI18Li}.
We then investigate the performance of each algorithm as a function of the available number of labeled nodes. 

\subsection{Datasets and Experimental Setup}
We test and compare the proposed methods on three comprehensive sets of real-world collaboration networks: {\tt Cora}, {\tt CiteSeer}, and {\tt Pubmed} provided by~\cite{sen2008collective}. Details of these three networks are given on Table~\ref{datatable}. For each dataset, nodes represent documents and edges represent the links among these documents, each document is defined by a bag-of-word feature vector, i.e., existence/non-existence of certain words are represented as $1/0$ values in this feature vector~\cite{AAAI18Li}.

For  \cotrain, we use the Python implementation provided by Li {\em et al.}~\cite{AAAI18Li}. 
We implement our algorithms in both Matlab, which is only used for offline computations, and Python. 
We estimate the required number of labeled nodes, $t$, using the lower
bound proposed by Li {\em et al.}~\cite{AAAI18Li}. 
These $t$ values are computed as $76$, $216$, and $975$ for respectively {\tt Cora}, {\tt CiteSeer}, and {\tt Pubmed}~\cite{AAAI18Li}. 

We assess the performance of the algorithms as a function of two key parameters: (i) $\eta$, the fraction of additional candidates considered for expansion and (ii) $K$ the number of clusters used to compute topological profiles. We consider five different values of $\eta$, namely $ \{ \frac{1}{10}t, \frac{3}{10}t, \frac{1}{2}t \frac{7}{10}t, t\}$. 
We consider three different values for $K$, namely $100$, $250$, and $500$, 
Finally, to assess the robustness of the proposed methods,
we fix $K$ and $\eta$ parameters that result in optimal performance for each algorithm and evaluate their performance as a function of the number of known labeled nodes. 

For the hyper-parameters of the GCNs, we follow Kipf and Welling's~\cite{Kipf} parameter settings. Namely, we use a learning rate of $0.01$, 200 maximum epochs, 0.5 dropout rate, $5\times 10^{-4}$ $L2$ regularization weight, 2 convulutional layers, and 16 hidden units~\cite{AAAI18Li}. For \textit{manifold learning } we use DPP method with its default settings~\cite{wachinger2015diverse}. For each experiment, we randomly divide the labels into small training sets, $0.1$\% training size for {\tt Cora} and {\tt CiteSeer}, and $0.02\%$ for the {\tt PubMed} dataset, and a set with 1000 samples for testing. Finally, we report the mean accuracy of 50 runs for each dataset for the first experiment and 10 runs for the second experiment.   All of the experiments are performed on a Dell PowerEdge T5100 server with two 2.4 GHz Intel Xeon E5530 processors and 32 GB of memory.


\subsection{Performance Evaluation}
We first compare the node classification performance of the three
methods using accuracy, number of correct prediction divided by total number of prediction~\cite{AAAI18Li}, as the performance criterion. 
The results of this analysis for three datasets are shown in Figure~\ref{fig:Performance}. 
As seen in the figure, on all three datasets, the GCN that uses labels expanded by one of the proposed algorithms delivers bes performance.
To be specific, on the Cora dataset, the accuracy of the GCN that uses
labels expanded by \tptrain\ clearly outperforms the GCN that uses labels
expanded by \cotrain.
As would be expected, the performance of label expansion by \tptrain\ is improved by the consideration of more candidate nodes for expansion (increasing $\eta$), but starts declining after a certain point (i.e,
consideration of too many candidates creates confusion). 
\tptrain\ outperforms \cotrain\ on the PubMed dataset as well, but the performance difference is less pronounced.
For the CiteSeer dataset, on the other hand, \mltrain\ delivers the best performance and the performance of \mltrain\ becomes more robust as
more communities are used to compute topological profiles.

We then investigate the performance of label expansion algorithms as a function of known labeled nodes. 
The result of this analysis are shown in Figure~\ref{fig:Performance2}. 
As seen in the figure, the accuracy provided by each of the three methods is improved consistently with the availability of more labeled samples.
It is also impressive for all the label expansion methods that the improvement
in accuracy appears to saturate when the number of labeled nodes reaches 32, suggesting that these methods truly provides GCNs with the opportunity to deliver their best performance with scarce training data.
Also impressively for the proposed method, either \tptrain\ or \mltrain\ drastically outperforms \cotrain\ when label shortage is at its worst, i.e., when only 2 labeled nodes are available for each class.
These results clearly demonstrate the effectiveness of topological similarity based algorithms in label expansion, suggesting that these algorithms have great potential in rendering GCNs useful even when training data is limited.Our proposed methods are robust when number of known labelled nodes are decreased. Moreover, the best performance gain is attained while known number of labelled nodes are small showing the value of using our approach while we have very limited labelled node set to train a GCN.


\section{Conclusions}
\label{sec:conclusion}

In this paper, we investigate the labeled node set expanding problem for training GCNs. To address this problem which is inherited by traditional deep learning, we present an alternate algorithm, \algo\, for extrapolating node labels in GCNs in the following three steps: first identifies communities in the graph, subsequently computes topological profiles for each node using their proximity to the communities, and finally assesses the {\em topological similarity} of each node to the nodes that are already labeled. 
It then expands the labels, for each class, by selecting the nodes that are most topologically similar to the nodes that are already labeled with that class. Using three large real-world networks that are commonly used in benchmarking GCNs, we systematically test the performance of the proposed algorithm and show that our approach outperforms existing methods for wide ranges of parameter values.

\bibliographystyle{spmpsci}      
\bibliography{mybib} 

\end{document}